\documentclass[10pt,twocolumn,letterpaper]{article}

\usepackage{wacv}
\usepackage{times}
\usepackage{epsfig}
\usepackage{graphicx}
\usepackage{amsmath}
\usepackage{amssymb}
\usepackage{booktabs}
\usepackage{mathtools}
\usepackage{makecell}
\usepackage{multirow}
\usepackage{bm}
\usepackage{bbm}
\usepackage{subcaption}

\def\wacvPaperID{1674} 

\wacvapplicationstrack 

\wacvfinalcopy 

\ifwacvfinal
\usepackage[breaklinks=true,bookmarks=false]{hyperref}
\else
\usepackage[pagebackref=true,breaklinks=true,colorlinks,bookmarks=false]{hyperref}
\fi

\begin{document}

\title{Multimodal contrastive learning for remote sensing tasks}

\author{Umangi Jain, Alex Wilson, Varun Gulshan\\
Google Research\\
{\tt\small \{jainumangi, alexwilson, varungulshan\}@google.com}
}

\maketitle
\thispagestyle{empty}

\begin{abstract}
   Self-supervised methods have shown tremendous success in the field of computer vision, including applications in remote sensing and medical imaging. Most popular contrastive-loss based methods like SimCLR, MoCo, MoCo-v2 use multiple views of the same image by applying contrived augmentations on the image to create positive pairs and contrast them with negative examples. Although these techniques work well, most of these techniques have been tuned on ImageNet (and similar computer vision datasets). While there have been some attempts to capture a richer set of deformations in the positive samples, in this work, we explore a promising alternative to generating positive examples for remote sensing data within the contrastive learning framework. Images captured from different sensors at the same location and nearby timestamps can be thought of as strongly augmented instances of the same scene, thus removing the need to explore and tune a set of hand crafted strong augmentations. In this paper, we propose a simple dual-encoder framework, which is pre-trained on a large unlabeled dataset ($\sim 1M$) of Sentinel-1 and Sentinel-2 image pairs. We test the embeddings on two remote sensing downstream tasks: flood segmentation and land cover mapping, and empirically show that embeddings learnt from this technique outperform the conventional technique of collecting positive examples via aggressive data augmentations.
\end{abstract}

\section{Introduction}

Recently, self-supervised learning (SSL) techniques have seen tremendous success as a way to pre-train supervised models. Popular self-supervised frameworks for computer vision tasks, including SimCLR~\cite{chen2020simple}, MoCo~\cite{he2020momentum}, MoCo-v2~\cite{chen2020improved}, Barlow twins~\cite{zbontar2021barlow}, BYOL~\cite{grill2020bootstrap}, learn representations by imposing invariance to several image augmentations. Many successful SSL techniques proposed in the past few years use a contrastive learning framework, where the pretext task is based on instance discrimination~\cite{wu2018unsupervised}, which treats every instance of an image as a separate class. The positive examples for each class (instance, in this case) are gathered by applying augmentations on each image. These augmentations are hand-crafted and the commonly used ones include random cropping, gaussian blurring, color jitter, and color drop. Some works argue that using only augmentations of the same image as positives can cause the model to learn only the most discriminative features in the image and the focus of the learning algorithm on shape, texture, and other specific properties in the image remains unclear~\cite{van2021revisiting}. This line of thought led some SSL methods to propose more elaborate means of collecting positive examples and capturing a richer set of deformations by using multi-crop (collecting multiple positive examples, each generated by applying augmentations on the local and global crops of the anchor image)~\cite{caron2020unsupervised}, clustering~\cite{wang2021unsupervised}~\cite{li2020prototypical}, nearest neighbors~\cite{van2021revisiting}, or using some additional metadata like geolocation for remote sensing pre-training~\cite{yamada2021geoclr}. 

These self-supervised techniques have also worked well for many remote sensing tasks. While ImageNet initialization is a strong baseline for remote-sensing applications like scene classification~\cite{pires2019convolutional}, pre-training on unlabeled satellite imagery using SSL techniques provides a further improvement, as shown by Patel \etal~\cite{patel2021evaluating} for flood segmentation, land cover mapping, and river segmentation tasks.

Another line of work explores SSL techniques specific to remote sensing domain. GeoCLR~\cite{yamada2021geoclr} generates positive pairs for contrastive learning by leveraging the geolocation metadata available in seafloor imagery and gathering images which are physically close as positives. However, this requires exploring the value of the distance within which an image is considered to be a positive instance. Ayush \etal~\cite{ayush2021geography} and Manas \etal~\cite{manas2021seasonal} propose in-domain contrastive learning based pre-training methods that leverage spatially aligned images over time to create temporal positive pairs. While these methods improve performance, they still rely on artificial augmentations with many hyper-parameters. One drawback of such pre-training methods is that any hyper-parameter tuning needed for building positive pairs is expensive -- there isn't a universal metric for measuring the quality of pre-trained embeddings~\cite{purushwalkam2020demystifying}, and any tuning needs to be done on a downstream task which make this parameter exploration slow and expensive.

Contrastive learning has also been applied to multimodal data as explored by methods such as CLIP~\cite{radford2021learning} and ALIGN~\cite{jia2021scaling} that apply contrastive loss between an image and its corresponding noisy text description on a large scale. The image and text dataset is scraped from the web by collecting images with a caption, and applying filters to clean the data. On this dataset, a dual-encoder model is trained which pushes embeddings of paired image and text closer in the feature space while pushing the non-matching instances away. In these settings, contrastive loss is closer to label-based classification with text modality acting as a noisy label for the image.  

Remote sensing offers another unique possibility of applying multimodal contrastive learning on images captured from different sources. Unlike optical RGB camera images (e.g., ImageNet~\cite{deng2009imagenet}, CoCo~\cite{chen2015microsoft} dataset), images obtained from different satellite constellations acquire different types of information about the scene:  Multispectral, LiDAR, hyperspectral, Synthetic Aperture Radar (SAR), all capture surface information differently and contain complementary information. Sheehan \etal~\cite{sheehan2019predicting} use satellite imagery with geo-tagged Wikipedia article pairs but it does not involve joint pre-training and the pair is directly used in the downstream task.  Heidler \etal~\cite{heidler2021self} propose a framework that exploits the correspondence between Sentinel-2 images with geo-tagged audio recordings for pre-training using batch-wise triplet loss and the model is fine-tuned for aerial image classification, aerial image segmentation, and audiovisual scene classification. 
Multimodal/multiview self-supervised contrastive learning has also been explored in~\cite{chen2021selfsup}~\cite{chen2021self}~\cite{saha2021self}~\cite{cha2021contrastive}~\cite{stojnic2021self}. However, the scale of pre-training in these works is small, and often the learned representations are not general purpose and limited to a single application. Jain \etal~\cite{jain2021multi}~\cite{jain2022self} also use contrastive learning based self-supervised approach with paired electro-optics and SAR imagery. 

We propose a multimodal framework for learning representations by using data from two different remote sensing satellites. Our hypothesis is that images from different remote sensors, captured at the same geolocation and close by timestamps, provide better positive examples for contrastive learning than what is obtained using the hand-crafted augmentation techniques. It also allows each modality to learn features which are more clearly visible/discriminable in the other modality. Images captured from different remote sensing sensors could be thought of as naturally occurring strong augmentations of the same scene. We use these naturally occurring augmentations to replace the synthetic augmentations of SimCLR and do a thorough comparison on two remote sensing datasets. Our work is closest to~\cite{jain2022self}, however, their work uses a smaller unlabeled dataset of $\sim90k$ samples and they pre-train by randomly choosing either a single band or three bands from both the modalities. Both the modalities share the same weights in their framework. Unlike~\cite{jain2022self}, we pre-train a dual-encoder model using contrastive loss between Sentinel-1 and Sentinel-2 imagery  (active radar and optical imagery, respectively; more details in Section \ref{Multi-satellite unlabeled dataset})  on $\sim1M$ data points and show the improvement in performance for both the modalities. Our main contributions in this work include:

\begin{itemize}
    \item We propose a multimodal framework for remote sensing applications which leverages the naturally occurring augmentations obtained from different remote sensors capturing the same scene.
    \item We do a large scale pre-training on $\sim1M$ Sentinel-1 and Sentinel-2 image pairs. 
    \item We test the quality of these embeddings on two publicly available downstream tasks of semantic segmentation: flood segmentation and land cover mapping.
\end{itemize}

\begin{figure*}[t]
\begin{center}
   \includegraphics[width=0.2\linewidth]{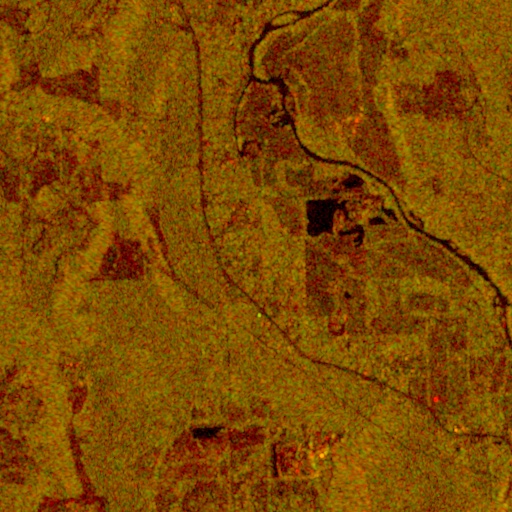}
   \includegraphics[width=0.2\linewidth]{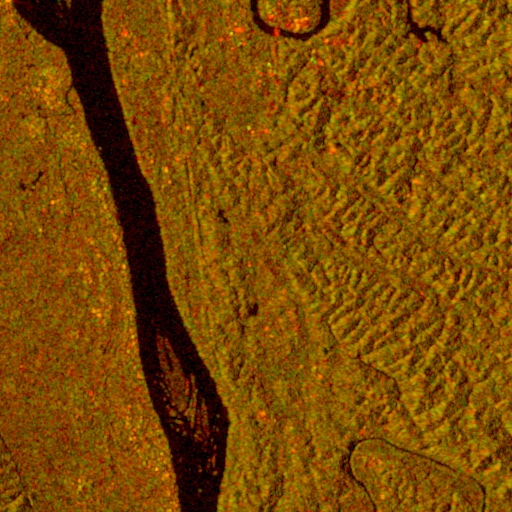}
   \includegraphics[width=0.2\linewidth]{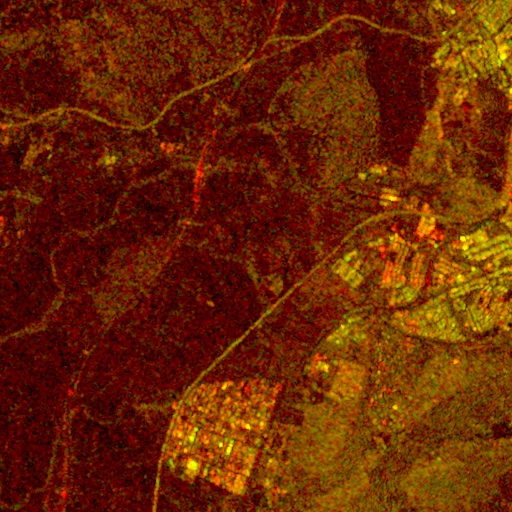}
   \includegraphics[width=0.2\linewidth]{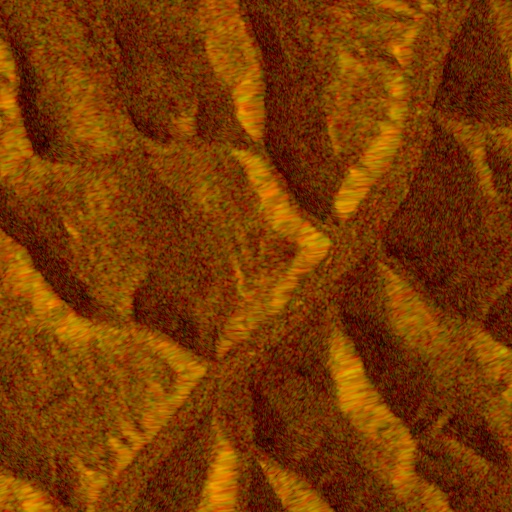}
   \includegraphics[width=0.2\linewidth]{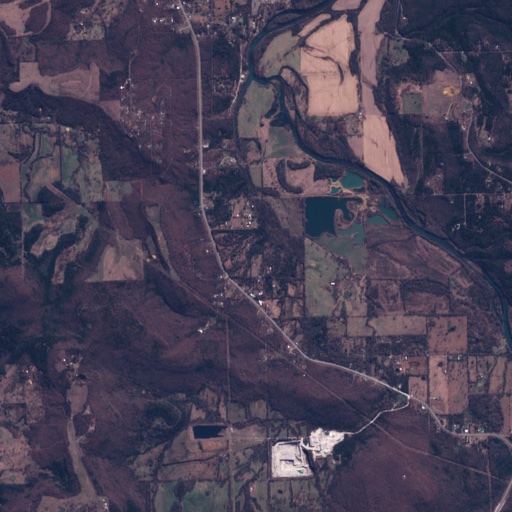}
   \includegraphics[width=0.2\linewidth]{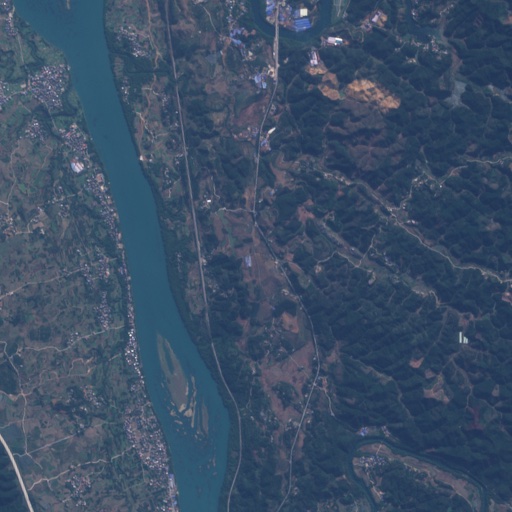}
   \includegraphics[width=0.2\linewidth]{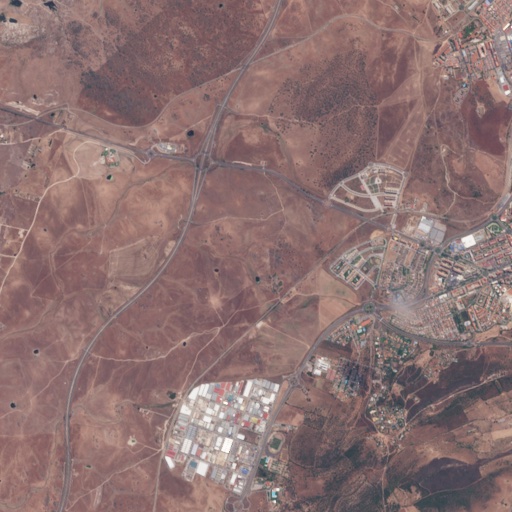}
   \includegraphics[width=0.2\linewidth]{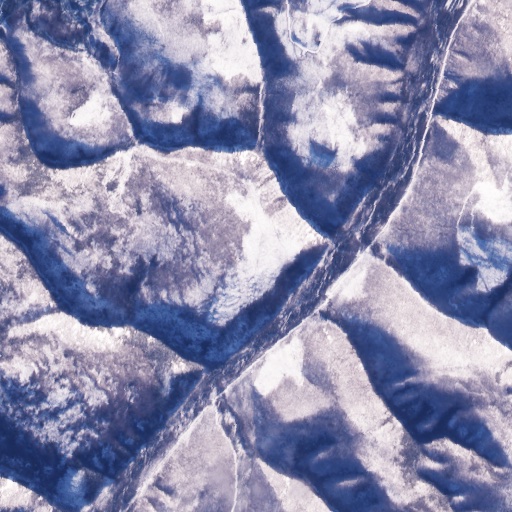}
\end{center}
\vspace{-1mm}
   \caption{Randomly sampled Sentinel-1 (top) and Sentinel-2 (bottom) image pairs sampled from the same location and close by timestamp. An additional channel of zeros is concatenated to the Sentinel-1 bands for visualization. }
\label{fig:unlabeled_data_samples}
\end{figure*}

\section{Multi-satellite unlabeled dataset}\label{Multi-satellite unlabeled dataset}

We extract paired Sentinel-1 and Sentinel-2 unlabeled pre-training data using Google Earth Engine~\cite{gorelick2017google}. 

\textit{Sentinel-1}: Sentinel-1~\cite{eesen1}~\cite{torres2012gmes} satellite constellation provides data from its Synthetic Aperture Radar (SAR) instrument, which is an active data collection sensor. It emits microwave radiation in the C-band (5.4GHz) which gets reflected from Earth after interacting with the surface and the bounced signal is recorded to characterize the surface properties. The images are pre-processed by thermal noise removal, radiometric calibration, and terrain correction and the pixel values of the exported images are in decibels. We collect Vertical Transmit-Vertical Receive (VV) and Vertical Transmit-Horizontal Receive (VH) bands from Sentinel-1 at 10m resolution.

\textit{Sentinel-2}: Sentinel-2~\cite{eesen2}~\cite{drusch2012sentinel} is a satellite constellation that acquires multispectral images at high resolution. It works passively by collecting light reflected from the surface of the Earth. We use Sentinel-2 Level 1C product which represents Top of Atmosphere (TOA) reflectance values. There are 13 spectral bands in this constellation, out of which we only use the RGB bands (B4, B3, and B2 respectively) captured at 10m resolution. 

The mechanism through which the two sources acquire imagery is very different, as visualized in Figure \ref{fig:unlabeled_data_samples}. SAR imagery captures signals irrespective of the weather conditions, clouds, or darkness, as opposed to optical imagery which shows high variation depending upon prevailing cloud cover. However, unlike optical imagery, SAR images are not as easy to interpret for non-expert humans and doesn't discriminate well between certain land cover types compared to optical imagery.

\subsection{Data sampling}
\begin{figure}
\begin{center}
   \includegraphics[width=0.8\linewidth]{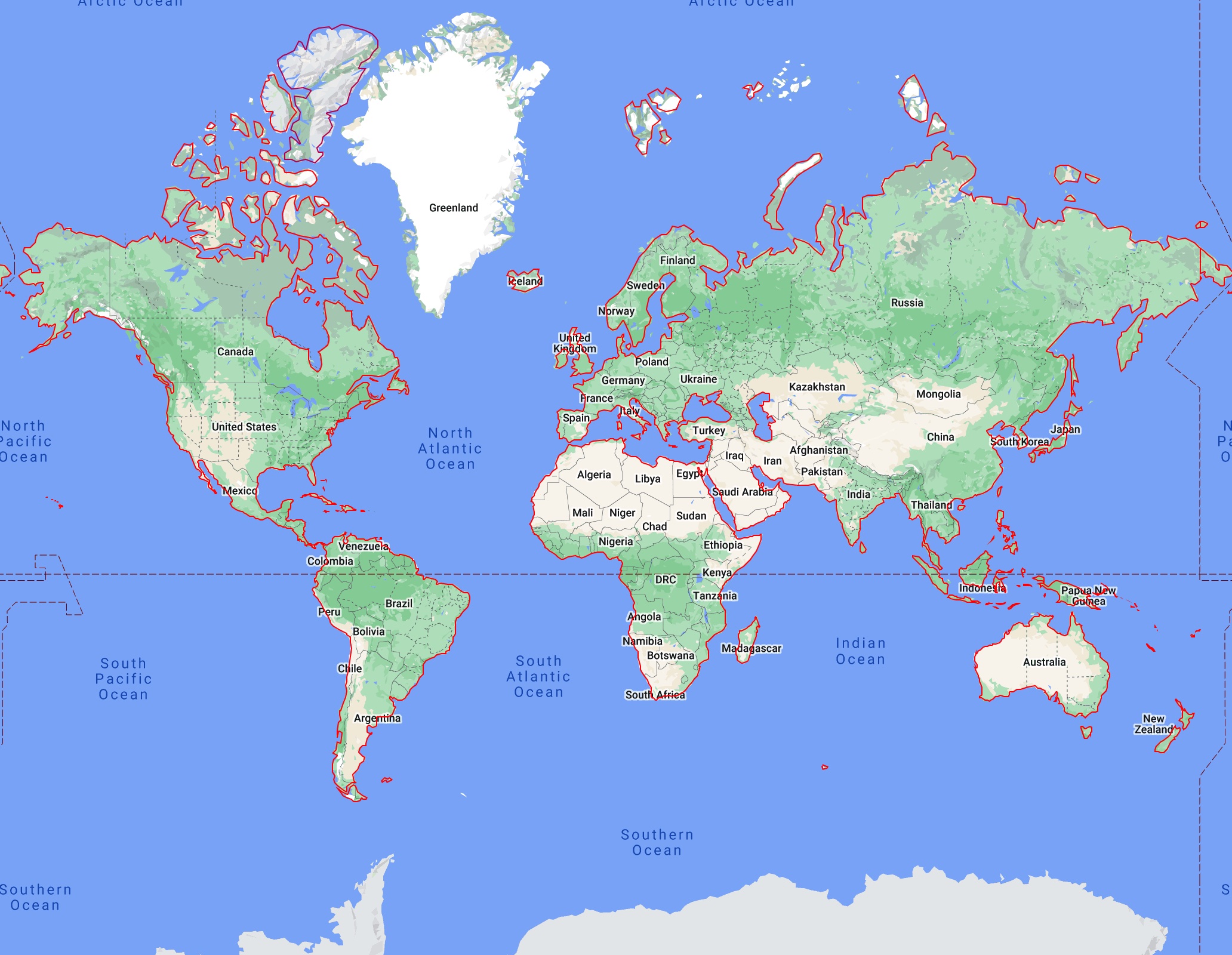}
\end{center}
\vspace{-1mm}
   \caption{Regions highlighted inside the red polygons are used to generate random lat-lon values, which are used to export unlabeled Sentinel-1 and Sentinel-2 paired imagery. }
\label{fig:global_landcover}
\end{figure}

We obtain Sentinel-1 and Sentinel-2 image pairs by generating IID samples of latitude, longitude from the global land mass, and IID samples of timestamp values collected over a period of 5 years from 31st December, 2016 to 31st December, 2021. We exclude Greenland and Antarctica from the global landmass as it might be difficult for contrastive loss to discriminate between homogenous images. An image pair is collected if there is a Sentinel-1 and a Sentinel-2 image available at the specified location and is within 30 days (in the past) of the specified timestamp. In cases where there are multiple images in the 30 days window, we choose the image which is closest to the specified timestamp. We apply cloud filtering to remove images with more than 15\% cloud coverage in Sentinel-2. This is done because cloud covered images obscure semantic information and do not contain features needed for learning in a contrastive learning framework. Sentinel-1 images are, however, not affected by cloud cover and do not need this filter. The total number of images collected for pre-training are 1,087,502 with an image size of 512 $\times$ 512 $\times$ 2 for Sentinel-1 and 512 $\times$ 512 $\times$ 3 for Sentinel-2. The regions from where the data is sampled are highlighted in Figure \ref{fig:global_landcover}.

\subsection{Image normalization}\label{Image normalization}

\begin{table*}
  \begin{center}
    {\small{
\begin{tabular}{l p{4cm} p{4cm} p{4cm}}
\toprule
 & Sen1Floods11 & Dynamic World (Sentinel-1) &  Dynamic World (Sentinel-2) \\
\midrule
Source & Sentinel-1 & Sentinel-1 & Sentinel-2\\
Bands & VV and VH & VV and VH & B4, B3, and B2 (RGB)\\
Resolution & 10m & 10m & 10m\\
Image size & 512x512 & 510x510 & 510x510 \\
Label classes & 2 (water, no water) & 9 (water, trees, grass, flooded vegetation, crops, shrub and scrub, built area, bare ground, snow and ice) &9 (water, trees, grass, flooded vegetation, crops, shrub and scrub, built area, bare ground, snow and ice) \\
No. train images & 252 & 18,293 &18,293 \\
No. validation images & 89 & 4,598 & 4,598 \\
No. test images & 90 & 407 &407 \\
Train regions & 11 flooding events from 6 continents & Global & Global \\
\bottomrule
\end{tabular}
}}
\end{center}
\vspace{-1mm}
\caption{Summary of key attributes for both the datasets used for evaluation.}
\label{tab:dataset_attributes}
\end{table*}

Data from both the satellite sources is normalized into a consistent range during data pre-processing. As discussed above, raw Sentinel-1 images are in decibels (dB) scale. We clip these images to a fixed range ([-20dB, 5dB]) and scale it linearly to pixel values between [0, 255]. Sentinel-2 images represent scaled TOA reflectance values. For Sentinel-2,  we use a logarithm-based nonlinear scaling method as in~\cite{brown2022dynamic}. This is done because cloudy pixels in Sentinel-2 have large reflectance values compared to the non-cloudy pixels and a linear scaling would result in a smaller range for the non-cloudy pixels.
\section{Downstream labeled datasets}
We evaluate trained embeddings on two publicly available labeled datasets: Sen1Floods11 and Dynamic World. 

\textit{Sen1Floods11}: We use the Sen1Floods11 dataset~\cite{bonafilia2020sen1floods11} for flood segmentation, which consists of labeled SAR images of flood scenes from 11 flooding events across 6 continents. The segmentation task in this dataset is to demarcate flooded regions using Sentinel-1 images. The authors released 4,831 images which were labeled using simple thresholding models, yielding noisy weak labels. A small subset of 446 images were hand corrected by experts, and we use only this subset for our experiments to avoid training data quality issues, and also test the effectiveness of representation learning in a data scarce setting.  The authors provide an IID partition of the data comprising of 252 train images, 89 validation images, and 90 test images. 

\textit{Dynamic World}: The segmentation task is to label each pixel with its land cover class in the publicly released Dynamic World dataset~\cite{brown2022dynamic}. We use the train and test dataset that has been labeled by human annotators. The number of Sentinel-2 training examples in the publicly available dataset is 22,906 and 409 samples in the test set. Sentinel-2 images in the Dynamic World dataset consist of 9 spectral bands. We augment this dataset by joining every Sentinel-2 image with a corresponding  Sentinel-1 image (VV and VH bands). Augmenting the dataset with Sentinel-1 images is useful as it allows for models to be trained and evaluated on SAR images that are robust to changing weather and lighting conditions. The joining criteria used is that a Sentinel-1 image should be available at the exact same location and within 3 months of the Sentinel-2 image. If this criteria cannot be met, we discard that example. Upon the join with Sentinel-1 images, 22,891 train images and 407 test images are obtained. We create an IID split of the train dataset into roughly 80:20 train and validation samples (as a separate validation set is not provided explicitly in the data).

\begin{figure*}[t]
\begin{center}
   \includegraphics[width=0.8\linewidth]{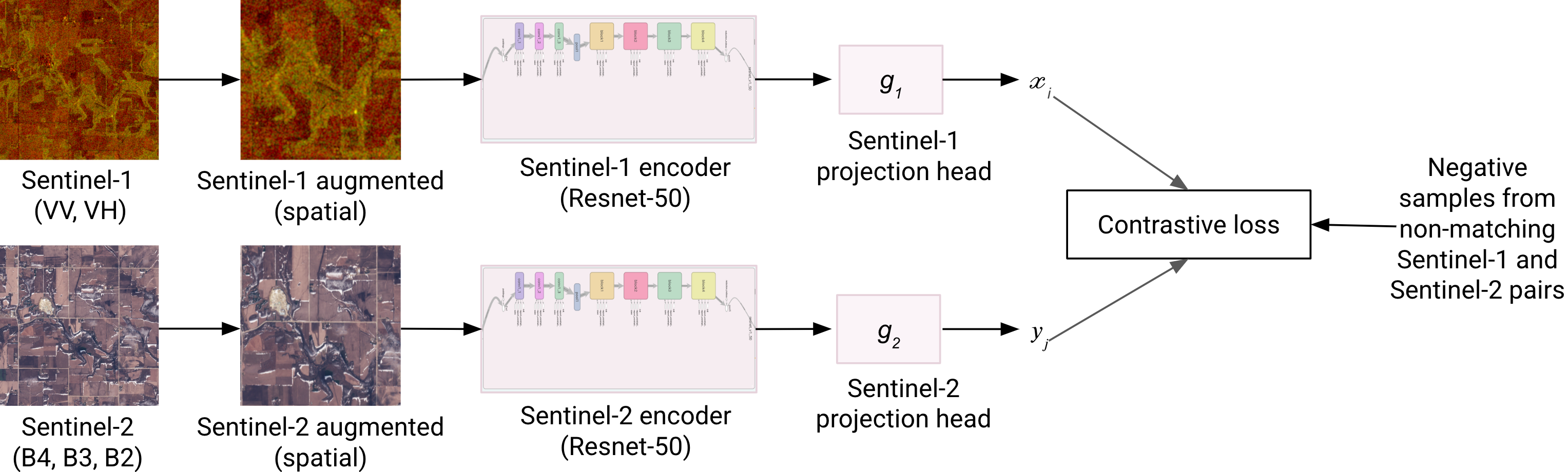}
\end{center}
   \caption{Overview of our multimodal contrastive learning framework. Sentinel-1 and Sentinel-2 are jointly mapped in the same space. The representations can be used for both Sentinel-1 or Sentinel-2 based downstream tasks. }
\label{fig:mm_framework}
\end{figure*}

This dataset is used to set up two downstream tasks, one that uses Sentinel-1 images only as inputs and the other only Sentinel-2 images. For the Sentinel-2 images, we only use the RGB bands as inputs. These two tasks are referred to as \textit{Dynamic World (Sentinel-1)} and \textit{Dynamic World (Sentinel-2)}, respectively, from here. 

The same image normalization that is applied on the unlabeled pre-training datasets (Section \ref{Image normalization}) is also used for all the downstream labeled datasets. Table \ref{tab:dataset_attributes} summarises the key attributes of these downstream labeled datasets. 

\section{Methods}
This section describes the multimodal contrastive learning framework, the baselines against which performance of the model is compared, and the overall experiment design.

\subsection{Multimodal pre-training}

We adapt the SimCLR contrastive learning framework for multimodal pre-training by constructing positive pairs from different satellite collections. The modalities we use come from an active and a passive remote sensor and act as natural augmentations to each other.  We take a pair of such images and apply spatial augmentations to them independently by taking a crop and resizing. Applying spatial augmentations is required to avoid the network from just learning the local edge features. No other augmentations (color drop, color jitter, or gaussian blur) are applied to these images, as these images are acquired from different sensors, hence capturing different features. 

We test our framework with the InfoNCE loss~\cite{oord2018representation} on a large batch size, similar to SimCLR. A spatially augmented pair of Sentinel-1 and Sentinel-2 image is passed through a dual-encoder architecture and a contrastive loss is applied to map their embeddings closer in feature space and away from the non-matching pairs. For a particular pair, both Sentinel-1 and Sentinel-2 images from the other pairs in the batch act as negative examples. Figure \ref{fig:mm_framework} illustrates our proposed method. 

The multimodal contrastive loss is defined per batch of images. Consider a batch of $N$ such image pairs, with $N$ images coming from Sentinel-1 $\{s_{1k}, k \in [1,...,N]\}$ and another $N$ from Sentinel-2 $\{s_{2k}, k \in [1,...,N]\}$. Each  $s_{1i}$ has a corresponding positive example $s_{2i}$  and the remaining   $2(N-1)$ images are considered as negative examples for this pair. Sentinel-1 images are encoded with an encoder network $f_1(.)$ and Sentinel-2 images are encoded with $f_2(.)$ to generate feature representations $h_{1i}$ and $h_{2i}$, respectively. These feature representations are passed through a non-linear projection head, $g_1(.)$ and $g_2(.)$ to produce embeddings $x_i$ and $y_i$, respectively (with $x_i = g_1(f_1(s_{1i}))$ and $y_i = g_2(f_2(s_{2i}))$). For a positive pair $i$, we define multimodal contrastive loss per batch as:
\begin{equation}
    l_i = l_{ixy} + l_{iyx}
\end{equation}
where $l_{ixy}$ and $l_{iyx}$ are defined as:
\begin{equation}
    l_{ixy} = -log\frac{exp(sim(x_{i}, y_{i})/\tau)}{
    \splitfrac{\Sigma_{k=1}^{N}\mathbbm{1}_{[k \neq i]}exp(sim(x_i, x_k)/\tau) +}{\Sigma_{k=1}^{N}exp(sim(x_i, y_k)/\tau)}} 
\end{equation}
\begin{equation}
    l_{iyx} = -log\frac{exp(sim(y_{i}, x_{i})/\tau)}{
    \splitfrac{\Sigma_{k=1}^{N}\mathbbm{1}_{[k \neq i]}exp(sim(y_i, y_k)/\tau) +} {\Sigma_{k=1}^{N}exp(sim(y_i, x_k)/\tau)}}
\end{equation}

$sim(.)$ denotes the cosine similarity function between two normalized vector embeddings. Like in SimCLR, the loss function causes matching image pairs to be drawn together, while pushing away all the other non-matching pairs in the batch. The same encoder networks $f_1(.)$ and $f_2(.)$ are also used as the encoder networks for the downstream tasks to allow transfer of these learned representations (more details in Section \ref{Training details}).

\subsection{Baselines}\label{Baselines}

We compare multimodal pre-trained models by fine-tuning on downstream tasks with the following baselines:

\textit{Random}: Supervised training of the downstream task from random initialization of the weights.

\textit{ImageNet}: Supervised fine-tuning from ImageNet pre-trained weights. While ImageNet pre-training is a strong baseline for consumer-camera-based vision problems, it is also used for remote-sensing applications~\cite{pires2019convolutional}~\cite{patel2021evaluating}. Since Sentinel-1 images have only 2 channels, we initialize the first convolutional layer of the encoder network with the average of the three channels.

\begin{table*}
  \begin{center}
    {\small{
\begin{tabular}{l p{4cm} p{4cm} p{4cm}}
\toprule
 & Sen1Floods11 & Dynamic World (Sentinel-1) &  Dynamic World (Sentinel-2) \\
\midrule
Encoder & ResNet-50 & ResNet-50 & ResNet-50\\
Evaluation Metric & Mean IoU of Water class & Classification accuracy & Classification accuracy\\
Number of train steps & 20,000 & 100,000 & 100,000\\
Sub-sampling experiment & \makecell[cl]{3 sets of 100\% \\ 5 sets of 10\%}  & \makecell[cl]{3 sets of 100\% \\ 5 sets of 10\% \\ 5 sets of 1\%}  & \makecell[cl]{3 sets of 100\% \\ 5 sets of 10\% \\ 5 sets of 1\%} \\
\bottomrule
\end{tabular}
}}
\end{center}
\caption{Training details of fine-tuning on the downstream datasets.}
\label{tab:finetuning_details}
\end{table*}
\textit{SimCLR}: We use SimCLR, a contrastive learning pre-training technique proposed in~\cite{chen2020simple}, as a baseline.  Chen \etal~\cite{chen2020simple} emphasized on using strong color augmentations by arguing that crops from the same image will have similar color distribution. Cole \etal~\cite{cole2022does} show that these augmentations are tuned for ImageNet and might not be optimal for other datasets. Therefore, we use the augmentation proposed in~\cite{patel2021evaluating} which is focused on remote sensing applications. The augmentations applied are distorted bounding box crop, random horizontal and vertical flips, color jitter, color drop (not applied on Sentinel-1 images), and random gaussian blur. The strength and probability for these augmentations are kept the same as in~\cite{patel2021evaluating}, which are different for Sentinel-1 and Sentinel-2 images. We pre-train two models for SimCLR, one pre-trained on each modality (Sentinel-1 and Sentinel-2). The dataset used is multi-satellite unlabeled data, described in Section \ref{Multi-satellite unlabeled dataset}, with each modality using only the data from its satellite constellation. We further improve the augmentations on Sentinel-1 and propose a stronger SimCLR baseline. The augmentations and results for that are reported in the Section \ref{Other experiments}.

\section{Training details}\label{Training details}

We compare the performance of the multimodal pre-trained models against all the baselines mentioned in Section \ref{Baselines} for each dataset.

\textit{Pre-training:} Both SimCLR and Multimodal models are trained using the same hyperparameters, similar to the setup in~\cite{chen2020simple}. ResNet-50~\cite{he2016deep} is used as the encoder architecture, for both SimCLR as well as our dual-encoder Multimodal pre-training. The first 7x7 convolutional layer in the architecture is replaced with two 3x3 convolutional layers, as done in  DeepLabv3+~\cite{chen2018encoder}. We train on a batch size of 4096, weight decay of $10^{-4}$, using the Layer-wise Adaptive Rate Scaling (LARS) optimizer~\cite{you2017large} with momentum 0.9. The initial learning rate is set to 0.48 with a cosine learning rate decay schedule (we reduce the learning rate by a factor of 10 compared to the default SimCLR training as the default one was high for training on this dataset, resulting in NaNs during training). The models are trained on 256 $\times$ 256 crop sizes till 160k steps. The temperature value for contrastive loss is kept constant as 0.1. The output from ResNet-50 is passed through a projection head giving 128 dimensional embeddings which are normalized before passing to the loss function. For the Multimodal model, the image encoder of the modality that comprises the downstream task is transferred for fine-tuning. 

Additional tuning for both SimCLR and Multimodal could further increase performance of these models. However, exploring the hyper-parameter space is computationally expensive for large-scale datasets (it is more expensive for SimCLR which has more augmentation parameters compared to the multimodal setup). 

\textit{Fine-tuning:} We use Deeplabv3+ encoder-decoder architecture~\cite{chen2018encoder} for segmentation tasks, with the same encoder as used for pre-training. We optimize for the cross-entropy loss. For ImageNet, SimCLR, and Multimodal pre-training, the initialization is done only for the encoder, and the atrous convolution and decoder layers are trained from scratch for all models. The weights are optimized on the train split, hyperparameter and checkpoint selection is done on the validation split and the final evaluation is done on the test split. We use a batch size of 64 with 321 $\times$ 321 image size for training. Atrous rates are set to (3, 6, 9) and the weight decay is kept at $10^{-6}$ for all experiments. The optimizer used is momentum with the momentum parameter set to 0.9. Polynomial schedule is used for the learning rate, starting from an initial value and decaying till zero with power 0.9. We do a sweep over the learning rate values, starting from $10^{-1}$ to $10^{-4}$, varying by a factor of $10^{-1}$. 

For understanding how the performance trend varies with label scarcity and to understand the impact of transferred representations in such settings, we sub-sample the labeled dataset at 1\%, 10\%, and 100\%. We sample 5 sets of the training data at 1\% and 10\%. These samples are chosen only once, are non-overlapping, and fixed for all experiments. For 100\%, we repeat the experiment on the entire dataset thrice. We do not conduct the 1\% experiment for Sen1Floods11 dataset as it has only 252 training examples, and 1\% of it would amount to only two samples which is too few for fine-tuning. For each dataset, the learning rate sweep is run on the entire training dataset and the same learning rate is used for the sub-sampling experiments. A summary of the training details is given in Table \ref{tab:finetuning_details}.

\section{Experimental evaluation}
The performance on Sen1Floods11 dataset is reported in pixel-wise Intersection over Union (IoU) of the water class. For Dynamic World, the metric used is overall classification accuracy.  A quantitative analysis of all the fine-tuned models is presented in Table \ref{tab:results_test}. 
Our model performs better than SimCLR across both the tasks and on all the sub-sampling splits of 1\%, 10\%, and 100\% of the training data. The results on the validation set are reported in \hyperref[appendix-valresults]{Appendix A}.

\begin{table*}
  \begin{center}
    {\small{
\begin{tabular}{l l l l l l}
\toprule
Dataset & Checkpoint & Learning rate &  1\% split & 10\% split & 100\% split\\
\midrule
\multirow{4}{*}{\makecell[cl]{Sen1Floods11\\(IoU water)}} & Random & 0.1 & & $55.22 \pm 5.29$ & $66.42 \pm 0.25$\\
&ImageNet&0.001& & $54.33 \pm 2.84$ & $65.56 \pm 0.47$ \\
&SimCLR &0.001 & &$55.35 \pm 4.95$ & $66.40 \pm 0.21$\\
&Multimodal &0.01 && \textbf{$\bm{57.89 \pm 5.65}$} & \textbf{$\bm{68.71 \pm 0.29}$} \\
\hline
\multirow{4}{*}{\makecell[cl]{Dynamic World \\ (Sentinel-1)\\(Classification 
accuracy)}} & Random & 0.01 & $47.49 \pm 1.31$ & $55.81 \pm 0.91$ & $61.86 \pm 0.48$ \\
& ImageNet & 0.01 & $50.71 \pm 1.52$ & $59.00 \pm 0.43$ & $65.93 \pm 0.20$ \\
& SimCLR &0.001 &$58.49 \pm 0.42$ & $63.70 \pm 0.43$ & $67.45 \pm 0.18$\\
& Multimodal &0.001 & \textbf{$\bm{59.59 \pm 0.96}$} & \textbf{$\bm{64.73 \pm 0.39} $} & \textbf{$\bm{68.72 \pm 0.55}$} \\
\hline
\multirow{4}{*}{\makecell[cl]{Dynamic World \\ (Sentinel-2)\\(Classification 
accuracy)}} & Random & 0.01 & $49.96 \pm 0.66$ & $62.90 \pm 0.92$ & $71.32 \pm 0.26$ \\
& ImageNet & 0.001 & $56.71 \pm 1.45$ & $69.00 \pm 0.46$ & $73.02 \pm 0.32$ \\
& SimCLR &0.001 &$65.87 \pm 1.08$ & $71.67 \pm 0.65$ &  {$\bm{74.96 \pm 0.24}$}\\
& Multimodal &0.001 & \textbf{$\bm{68.07 \pm 1.02}$} & \textbf{$\bm{72.93 \pm 0.26} $} & $74.95 \pm 0.18$ \\
\bottomrule
\end{tabular}
}}
\end{center}
\vspace{-3mm}
\caption{Results on the test set of Sen1Floods11 and Dynamic World dataset. The numbers are aggregated mean and standard deviation of respective metrics.}
\label{tab:results_test}
\end{table*}


\subsection{Sen1Floods11}
Table \ref{tab:results_test} shows water IoU on the test split of Sen1Floods11 dataset. SimCLR shows comparable performance to training with random initialization on both 10\% and 100\% split. Despite being pre-trained on large datasets, both ImageNet and SimCLR do not significantly outperform the random initialization. Fine-tuning on multimodal pre-training results in the best IoU values of the water class, improving by +2.31\% over SimCLR for the 100\% split. 

As discussed in~\cite{he2019rethinking}, ImageNet pre-training can speed up convergence early in training, but does not necessarily provide regularization or improve final target task performance. Multimodal consistently outperforms ImageNet initialization, with +3.5\% and +3.1\% improvement in the absolute IoU value (for water class) on 10\% and 100\% of the training data, respectively, reinforcing the findings in~\cite{cole2022does}~\cite{patel2021evaluating} that in-domain pre-training benefits more than cross-domain pre-training. 

\subsection{Dynamic World}

\textit{Dynamic World (Sentinel-1)}: For land cover segmentation, ImageNet initialization provides huge gain over training from random initialization, improving the absolute classification accuracy by +3.2\% for the 1\% split. The performance is boosted further when fine-tuning from SimCLR initialization. Multimodal performs the best, providing a 1 to 1.3\% increase in the classification accuracy over SimCLR across the various sub-samples of the dataset.

The results on 1\% and 10\% experiments show that in-domain multimodal pre-training can give huge gains, providing +8.9\% and +5.7\% absolute classification accuracy improvement, respectively, over initialization from ImageNet, making the learning extremely data efficient. 

\textit{Dynamic World (Sentinel-2)}: We observe a similar trend when using Sentinel-2 images as inputs. ImageNet boosts the performance significantly over fine-tuning from scratch, providing a +6.8\% gain in absolute classification accuracy for the 1\% split.  Multimodal pre-training gives +2.2\% and +1.3\% absolute improvement on classification accuracy over SimCLR for 1\% and 10\% split, respectively. The performance seems to saturate when using all the labeled data, with SimCLR and multimodal performing almost equally well, closely followed by ImageNet.

The overall performance of land cover segmentation on Dynamic World (Sentinel-2) is higher than Dynamic World (Sentinel-1). For 100\% of the data, Sentinel-2 bands achieve +6.2\% higher absolute classification accuracy than Sentinel-1 after multimodal fine-tuning. This is because Sentinel-2 bands are more discriminative for land cover classes, as can be visualized in Figure \ref{fig:unlabeled_data_samples}.

\begin{figure*}[t]
\begin{center}
    \begin{subfigure}{0.49\textwidth}
         \centering
         \includegraphics[width=\linewidth]{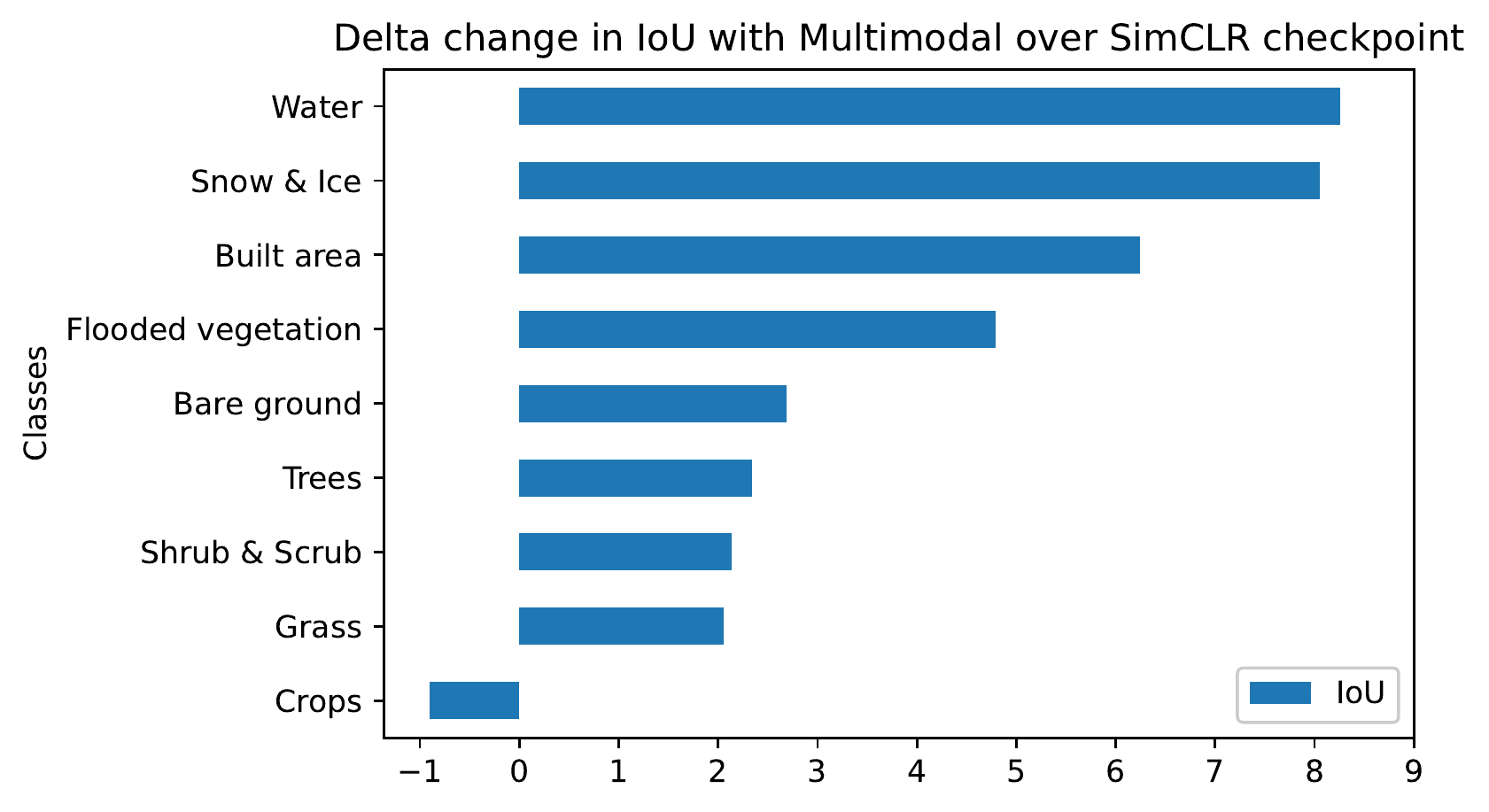}
         \caption{Dynamic World (Sentinel-2)}
         \label{fig:DW_S2_diff}
   \end{subfigure}
   \hfill
   \begin{subfigure}{0.49\textwidth}
         \centering
         \includegraphics[width=\linewidth]{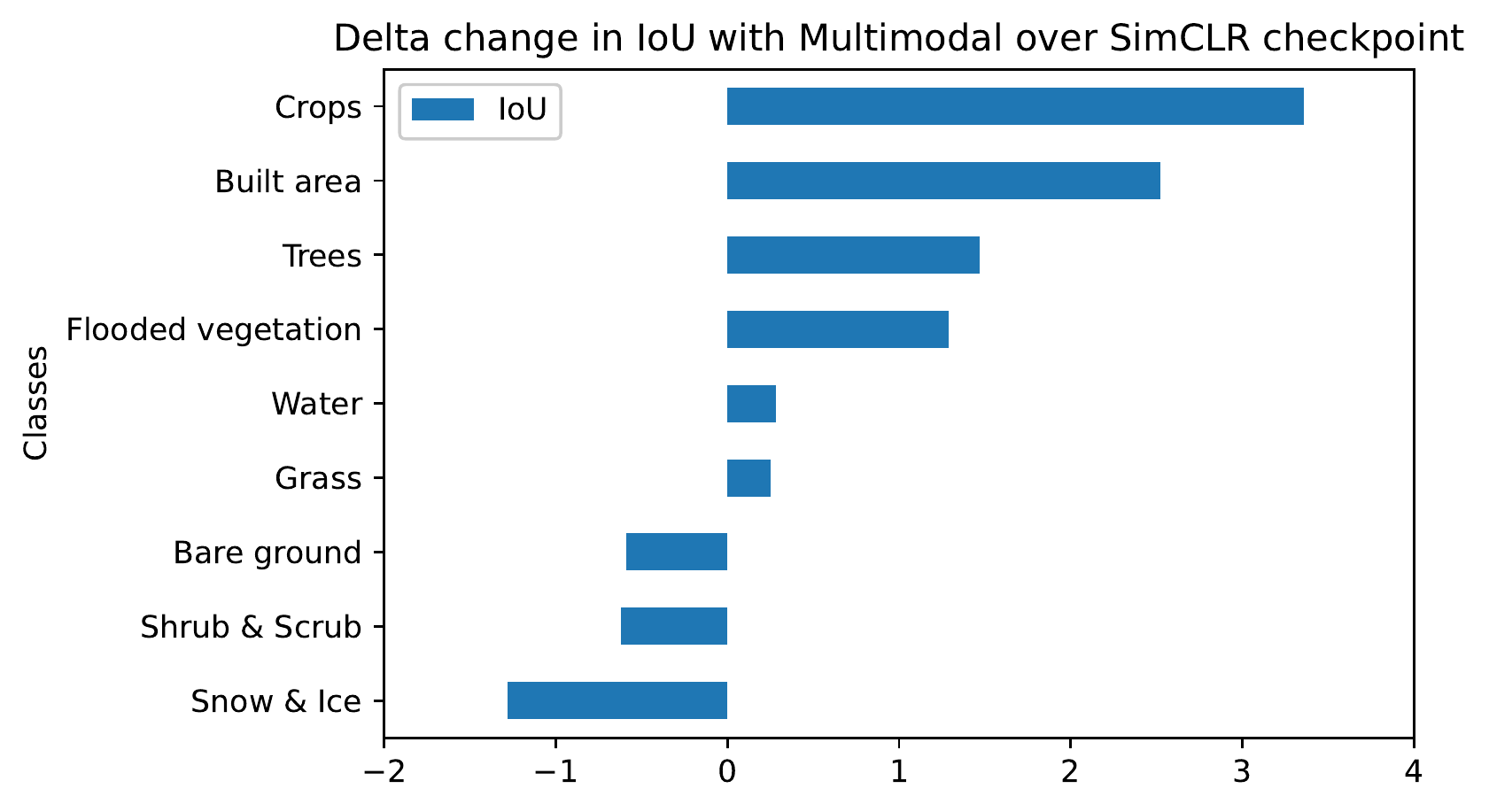}
         \caption{Dynamic World (Sentinel-1)}
         \label{fig:DW_S1_diff}
   \end{subfigure}
\end{center}
\vspace{-3mm}
   \caption{Average absolute IoU difference between models fine-tuned using Multimodal and SimCLR checkpoints for each class of Dynamic World on (a) Sentinel-2 (b) Sentinel-1.
}
\label{fig:per_class_miou}
\end{figure*}

\textbf{Learning complimentary information across modalities}: We compare the performance of our multimodal checkpoint with the SimCLR baseline for each class in the Dynamic World (Sentinel-1) and Dynamic World (Sentinel-2) datasets to understand how the performance of each class is affected when using multimodal checkpoints. To measure the performance for each class, pixel-wise IoU metric is used, and the reported result is the average IoU per class from the 5 sets of 1\% training data experiments. Our findings are summarized in Figure \ref{fig:per_class_miou}. 

We hypothesize that multimodal contrastive learning enables the network to learn complimentary information from different modalities, i.e., it allows each modality to learn features which are more clearly distinguishable in the other modality, while also retaining the features from the original modality.

Consider fine-tuning on Dynamic World (Sentinel-2) dataset, the multimodal checkpoint outperforms SimCLR checkpoint for 8 out of the 9 Dynamic World classes. Highest gain in the absolute IoU metric is observed in water class by +8.26\%, followed by snow and ice with a +8.05\% absolute IoU gain, and built area by +6.24\%. Crop segmentation shows a small degradation of 0.9\%.

Active sensors, like SAR, are known to be good at identifying very smooth surfaces (like calm water and smooth ice) and very rough surfaces (like man-made buildings)~\cite{anusha2020flood}~\cite{martinis2015backscatter}~\cite{dierking2013sea}~\cite{moreira2013synthetic}. In line with this, we observe that water, ice/snow, and built area classes observe maximum gains for Sentinel-2 imagery with multimodal pre-training. Compared to SimCLR pre-training on just Sentinel-2 images, multimodal training also incorporates the discriminative capabilities of Sentinel-1 during pre-training.

We observe a similar pattern when fine-tuning on Dynamic World (Sentinel-1). The multimodal initialization improves over SimCLR initialization in 6 out of the 9 classes. Crops and built area show improved absolute IoU by +3.36\% and +2.52\%, respectively. While the overall classification accuracy and mean IoU value is improved when fine-tuning on multimodal checkpoint, the per-class IoU for snow and ice, shrub and scrub, and bare ground deteriorate by 1.28\%, 0.62\%, 0.59\%, respectively. Optical imagery is the preferred data source for agricultural crop classification~\cite{mcnairn2009integration}, as multispectral optical imagery can measure and monitor the growth, stage type, and crop health.  Multimodal learning can leverage this discriminative characteristic of Sentinel-2 to enhance the Sentinel-1 embeddings.

These results align with our hypothesis that multimodal learning improves the representations for both the modalities, Sentinel-1 and Sentinel-2, as each modality also learns complimentary information from the other modality.

\subsection{Other experiments}\label{Other experiments}
\begin{itemize}
    \item \textbf{Improving SimCLR augmentations:} We explore further optimizations over the SAR image SimCLR augmentations used in~\cite{patel2021evaluating} (these include random flips, color jitter, and Gaussian blur to the image crops). We reduce the intensity and probability of applying color jitter and the other parameters are kept the same. With this improved pre-trained Sentinel-1 SimCLR model, we observe a +1.2\% and +1.4\% gain in absolute IoU for water class on 10\% and 100\% split for Sen1Floods11 dataset compared to the SimCLR checkpoint described in Section \ref{Baselines}. For the Dynamic World (Sentinel-1) dataset, the change in absolute classification accuracy for 1\%, 10\%, and 100\% split is -0.6\%, +0.4\%, +0.5\%, respectively.
    
    While this makes for a stronger baseline for SimCLR on Sentinel-1 images, our multimodal model still outperforms SimCLR on both the tasks. It shows that exploring optimal parameters for these hand chosen augmentations (operation, strength, and probability) is a computationally expensive task and influences the performance of SimCLR heavily. Our model eliminates the requirement of looking for optimal augmentation hyperparameters and leverages images from different sensors instead to provide a more effective set of augmentations. The exact augmentations and results are detailed in \hyperref[appendix-simclraug]{Appendix B}.

\item{\textbf{Training speed:}} We observe that the training speed of fine-tuning using Multimodal checkpoint is faster or comparable to SimCLR. Models fine-tuned on Multimodal checkpoint attain peak performance in $\sim2.4$x less steps compared to SimCLR on Sen1Floods11 and comparably on Dynamic World. The details of the training curves are illustrated in \hyperref[appendix-trainingspeed]{Appendix C}.
\end{itemize}

\section{Conclusion}
We present a simple method of leveraging abundant amounts of unlabeled pre-training data across different input modalities that remote sensing satellites offer. Our method avoids selecting and tuning hand-crafted augmentations for satellite images, and only requires simple spatial augmentations to work. Our dual-encoder multi-modal models, pre-trained on large multi-satellite unlabeled datasets using contrastive loss outperform the traditional baselines (ImageNet initialization, SimCLR) on two remote sensing tasks. A similar training architecture could also be applied to other modality combinations outside of the  two satellites we explored in this paper. Other input types, like temperature, precipitation, elevation, geo-tagged articles, or audio, all contain rich features that could improve representations across modalities  during pre-training and can be explored in future work.

\section{Acknowledgments}
We would like to thank Vishal Batchu for helping us generate multi-satellite unlabeled dataset, Shubhika Garg for insightful discussions on flood mapping experiments, Chaitanya Patel for assisting in baseline experiments, and John Platt and Rob von Behren for reviewing the paper and providing invaluable feedback.
{\small
\bibliographystyle{ieee_fullname}
\bibliography{egbib}
}

\newpage
\onecolumn
\appendix

\renewcommand{\thesection}{\Alph{section}}

\section{Results on validation set}\label{appendix-valresults}
\setcounter{table}{0}
\renewcommand{\thetable}{A\arabic{table}}
\setcounter{figure}{0}
\renewcommand{\thefigure}{A\arabic{figure}}
The hyper-parameter and checkpoint selection is done on the validation set of the labeled data for each dataset. The results are detailed in Table \ref{tab:results_val}.

\begin{table*}[h]
  \begin{center}
    {\small{
\begin{tabular}{l l l l l l}
\toprule
Dataset & Checkpoint & Learning rate &  1\% split & 10\% split & 100\% split\\
\midrule
\multirow{4}{*}{\makecell[cl]{Sen1Floods11\\(IoU water)}} & Random & 0.1 & & $51.04 \pm 5.38$ & $63.50 \pm 1.03$\\
&ImageNet&0.001& & $54.66 \pm 3.29$ & $64.21 \pm 1.66$ \\
&SimCLR &0.001 & &$51.24 \pm 4.35$ & $64.30 \pm 0.36$\\
&Multimodal &0.01 && \textbf{$\bm{55.83 \pm 3.5}$} & \textbf{$\bm{66.89 \pm 0.15}$} \\
\hline
\multirow{4}{*}{\makecell[cl]{Dynamic World \\ (Sentinel-1)\\(Classification 
accuracy)}} & Random & 0.01 & $51.67 \pm 0.88$ & $59.99 \pm 0.80$ & $65.39 \pm 0.47$ \\
& ImageNet & 0.01 & $55.02 \pm 0.95$ & $63.50 \pm 0.45$ & $69.32 \pm 0.16$ \\
& SimCLR &0.001 &$61.75 \pm 0.60$ & $67.30 \pm 0.25$ & $70.31 \pm 0.16$\\
& Multimodal &0.001 & \textbf{$\bm{63.26 \pm 0.61}$} & \textbf{$\bm{68.49 \pm 0.23} $} & \textbf{$\bm{71.26 \pm 0.06}$} \\
\hline
\multirow{4}{*}{\makecell[cl]{Dynamic World \\ (Sentinel-2)\\(Classification 
accuracy)}} & Random & 0.01 & $56.39 \pm 1.44$ & $67.56 \pm 0.47$ & $73.92 \pm 0.18$ \\
& ImageNet & 0.001 & $62.06 \pm 1.33$ & $71.88 \pm 0.36$ & $75.06 \pm 0.30$ \\
& SimCLR &0.001 &$69.06 \pm 1.30$ & $74.56 \pm 0.28$ &  {${77.00 \pm 0.13}$}\\
& Multimodal &0.001 & \textbf{$\bm{70.74 \pm 1.11}$} & \textbf{$\bm{74.98 \pm 0.18} $} & $\bm{77.14 \pm 0.17}$ \\
\bottomrule
\end{tabular}
}}
\end{center}
\caption{Results on the validation set of Sen1Floods11 and Dynamic World dataset. The numbers are aggregated mean and standard deviation of respective metrics.}
\label{tab:results_val}
\end{table*}


\section{Improved SimCLR augmentation}\label{appendix-simclraug}
\setcounter{table}{0}
\renewcommand{\thetable}{B\arabic{table}}
\setcounter{figure}{0}
\renewcommand{\thefigure}{B\arabic{figure}}
As discussed in the paper, we propose an improved set of augmentation for SAR images for SimCLR pre-training, compared to ~\cite{patel2021evaluating}. We reduce the color jitter strength and probability and all the other parameters are kept the same. The list of augmentations applied on SAR images include:
\begin{itemize}
\itemsep0em 
    \item Random horizontal flip with probability 0.5
    \item Random vertical flip with probability 0.5
    \item Color distortion (jitter) with a probability of 0.5 and strength 5
    \item Random Gaussian blur with a probability of 0.5 and strength 4
\end{itemize}

\begin{table*}[h]
  \begin{center}
    {\small{
\begin{tabular}{l l l l l l}
\toprule
Dataset & Checkpoint & Learning rate &  1\% split & 10\% split & 100\% split\\
\midrule
\multirow{3}{*}{\makecell[cl]{Sen1Floods11\\(IoU water)}} & SimCLR &0.001 & &$55.35 \pm 4.95$ & $66.40 \pm 0.21$\\
&SimCLR (improved augmentation) &0.01 & &$56.55 \pm 4.37$ & $67.83 \pm 0.42$\\
&Multimodal &0.01 && \textbf{$\bm{57.89 \pm 5.65}$} & \textbf{$\bm{68.71 \pm 0.29}$} \\
\hline
\multirow{3}{*}{\makecell[cl]{Dynamic world \\ (Sentinel-1)\\(Classification 
accuracy)}} & SimCLR &0.001 &$58.49 \pm 0.42$ & $63.70 \pm 0.43$ & $67.45 \pm 0.18$\\
& SimCLR (improved augmentation) &0.001 &$57.86 \pm 1.08$ & $64.14 \pm 0.07$ & $67.92 \pm 0.42$\\
& Multimodal &0.001 & \textbf{$\bm{59.59 \pm 0.96}$} & \textbf{$\bm{64.73 \pm 0.39} $} & \textbf{$\bm{68.72 \pm 0.55}$} \\
\bottomrule
\end{tabular}
}}
\end{center}
\caption{Results on the test set of Sen1Floods11 and Dynamic world (Sentinel-1) dataset with improved SAR augmentations.}
\label{tab:sar_lessjitter}
\end{table*}
More details about the definition and implementation of these operations can be found in~\cite{chen2020simple}. Table \ref{tab:sar_lessjitter} shows the results with this augmentation for Sen1Floods11 and Dynamic World (Sentinel-1) dataset. As outlined in the paper, fine-tuning with Multimodal checkpoint still outperforms SimCLR with improved SAR augmentation (by +0.88\% absolute IoU for Sen1Floods11 and +0.80\% absolute classification accuracy for Dynamic World (Sentinel-1) on the 100\% split). Similar trend follows for the other sub-sampling splits as well.

\section{Training Speed}\label{appendix-trainingspeed}
\setcounter{table}{0}
\renewcommand{\thetable}{C\arabic{table}}
\setcounter{figure}{0}
\renewcommand{\thefigure}{C\arabic{figure}}
We observe that fine-tuning with Multimodal pre-trained checkpoint reaches peak performance in nearly $\sim 2.4$x less steps (averaged over multiple runs) compared to SimCLR on Sen1Floods11 dataset and comparably on Dynamic World. Figure \ref{fig:training_time} compares validation curves for both Multimodal and SimCLR. All curves correspond to training with 100\% training data. 

\begin{figure*}[h]
\begin{center}
    \begin{subfigure}{0.45\textwidth}
         \centering
         \includegraphics[width=\linewidth]{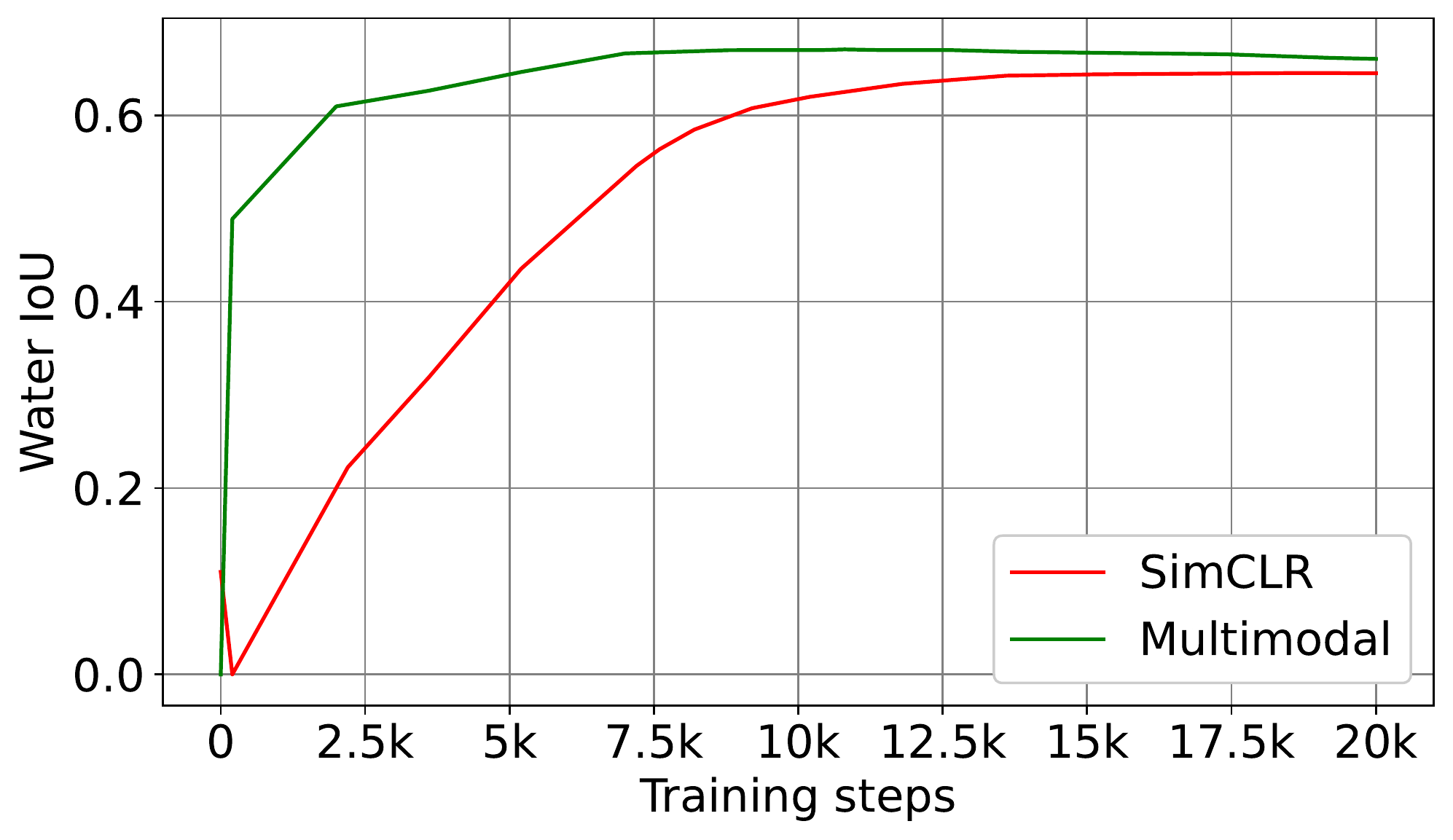}
         \caption{Sen1Floods11}
         \label{fig:A2_Sen1Floods11}
   \end{subfigure}
   \hfill
   \begin{subfigure}{0.45\textwidth}
         \centering
         \includegraphics[width=\linewidth]{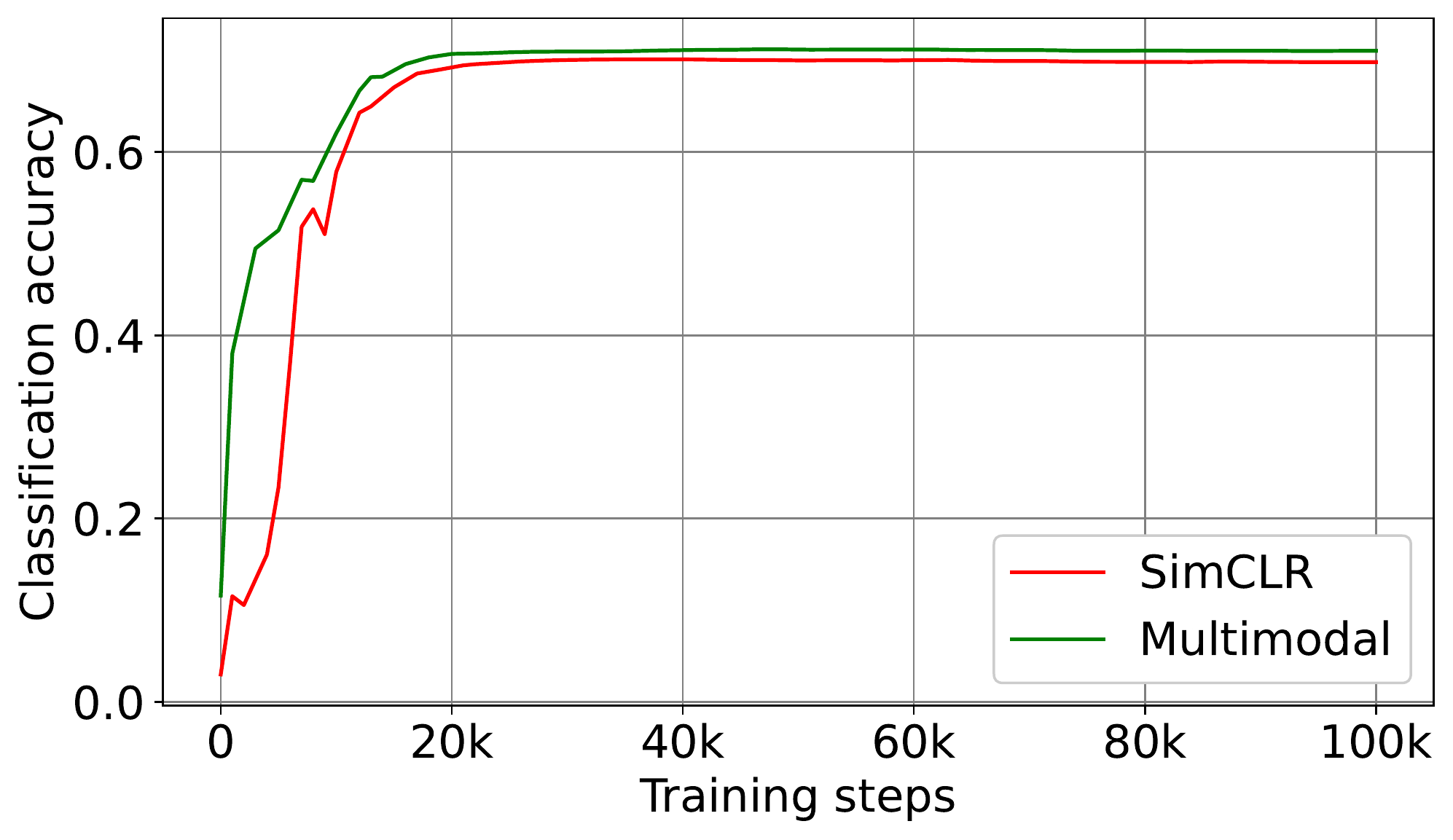}
         \caption{Dynamic World (Sentinel-1)}
         \label{fig:A2_dw_s1}
   \end{subfigure}
   \hfill
\end{center}
   \caption{Validation curves comparing fine-tuning using Multimodal and SimCLR pre-trained models on (a) Sen1Floods11 (b) Dynamic World (Sentinel-1) dataset.}
\label{fig:training_time}
\end{figure*}
\end{document}